\documentclass[runningheads]{llncs}
\usepackage{graphicx}
\usepackage{amsmath,amssymb} 
\usepackage{color}
\usepackage[width=122mm,left=12mm,paperwidth=146mm,height=193mm,top=12mm,paperheight=217mm]{geometry}

\usepackage{multirow}
\usepackage{array}
\usepackage{overpic}
\usepackage{rotating}
\usepackage{algorithm}
\usepackage{algorithmic}
\usepackage{float}
\usepackage{subfig}
\usepackage{authblk}
\begin{document}

\pagestyle{headings}
\mainmatter
\def\ECCV18SubNumber{***}  

\title{Effective Occlusion Handling for Fast Correlation Filter-based Trackers} 

\titlerunning{ECCV-18 submission ID \ECCV18SubNumber}

\authorrunning{ECCV-18 submission ID \ECCV18SubNumber}

\author{Zheng ZHANG\thanks{zhangz@cse.cuhk.edu.hk}}
\author{T.T. WONG\thanks{ttwong@cse.cuhk.edu.hk}}
\affil{ The Chinese University of Hong Kong}


\maketitle
\begin{abstract}
  
  Correlation filter-based trackers heavily suffer from the problem of multiple peaks in their response maps incurred by occlusions. Moreover, the whole tracking pipeline may break down due to the uncertainties brought by shifting among peaks, which will further lead to the degraded correlation filter model. To alleviate the drift problem caused by occlusions, we propose a novel scheme to choose the specific filter model according to different scenarios. Specifically, an effective measurement function is designed to evaluate the quality of filter response. A sophisticated strategy is employed to judge whether occlusions occur, and then decide how to update the filter models. In addition, we take advantage of both log-polar method and pyramid-like approach to estimate the best scale of the target. We evaluate our proposed approach on VOT2018 challenge and OTB100 dataset, whose experimental result shows that the proposed tracker achieves the promising performance compared against the state-of-the-art trackers.


\end{abstract}

\section{Introduction}
Visual tracking is one of the most important research topics in computer vision. It is widely used in various applications including video surveillance, robotics, driverless vehicles and human-computer interaction. Although great progress has been made in past decades, many problems remain difficult to be solved. For example, tracking object having occlusions is one of the typical challenge. Even with the powerful deep learning features, the state-of-the-art trackers usually fail in such situations. 

The reason is that the appearance of tracked target is interrupted by occlusions, which makes the visual object tracker to drift into the background. Specifically, the response map for a correlation filter-based tracker deviates from one-peak distribution which should be a Gaussian-like shape.
This makes the output of correlation filter-based tracker ambiguous that breaks down online learning pipeline in updating filter model. It eventually leads tracking failure. To deal with such issue of model degradation,~\cite{lukevzivc2017fclt} proposes an effective updating strategy to avoid the learned filter model being contaminated by occlusion.
However, their measurement function to determine whether occlusion happens is so far quite simple, which obtains the inferior performance.
Meanwhile, ~\cite{bhat2018unveiling} proposes a measurement function to compute the quality of response map, which is promising practically.
Although considering both the accuracy and robustness of the response map, this measurement function does not take into account of the non-linearity mismatching between the numerator and denominator in the measurement function, where the non-linearity of denominator is larger than that of numerator.

To address the above limitations, we propose an effective occlusion handling scheme for correlation filter-based trackers in this paper. Specifically, we employ an extra correlation filter tracker to locate the object while occlusion occurs as in~\cite{lukevzivc2017fclt,bhat2018unveiling}. Thus, the key of our proposed approach is to find an effective measurement function to judge whether occlusion occurs in video sequence. To this end, we propose a proportional function to measure the quality of filter response map. Instead of just using a simple value of quality of response map as the threshold~\cite{5539960}, our presented method captures the sudden quality drop of the response map to trigger the occlusion handling routines. Our proposed scheme increases the non-linearity of numerator in measurement function for the filter response map~\cite{bhat2018unveiling} by increasing its power degree. Therefore our quality measurement function is capable of trade-offing between accuracy and robustness of the response.  To deal with scale variations, we take advantage of both the log-polar method~\cite{DBLP:journals/corr/abs-1712-05231} and pyramid-like approach~\cite{danelljan2014accurate} to estimate the best scale of target. We have conducted comprehensive experiments on both VOT-2018 challenge~\cite{VOT_TPAMI} and OTB-100 benchmark~\cite{wu2015object}, whose promising results demonstrate the efficacy of our proposed tracker comparing against the state-of-the-art trackers.

The main contributions of this paper can be summarized as follows: 1) a novel scheme with sophisticated decision policy to deal with occlusions, in which two correlation filter models are employed to compute the position of target alternatively. Note that our proposed framework can be applied to other correlation filter-based tracking methods; 2) a measurement function to evaluate the quality of response map that help to detect occlusions by capturing the fast drop of the quality value; 
3) the promising results on several benchmarks, especially on the VOT-2018 real-time sub-challenge and main challenge~\cite{VOT_TPAMI}.

 \section{Related work}

  {}
 
 During past decades, intensive research efforts \cite{yilmaz2006object} have been devoted to fast visual object tracking. In this paper, we focus our attention on the correlation filter-based tracker, which is a promising approach currently. 
 
 Firstly, we investigate the correlation filter-based approaches with realtime performance. The pioneering MOSSE tracker~\cite{5539960} runs about 800 fps as a result of applying the Fourier domain to speed up tracking. However, it obtains inferior performance due to the lack of effective sampling scheme. To this end, KCF~\cite{henriques2015high} employs kernel tricks and circular matrix into correlation tracker and achieves very large margin of performance improvements. Currently, most of real-time correlation filter-based trackers are based on KCF~\cite{henriques2015high}.
 DSST~\cite{danelljan2014accurate} learns correlation filters from a scale pyramid samples.
 LDES~\cite{DBLP:journals/corr/abs-1712-05231} solves scale change and rotation of target by introducing log-polar into correlation filter framework.
 BACF~\cite{galoogahi2017learning}  shifts the search region and generates all circular shifts of the frame to learn from all possible positive and negative samples extracted from the entire frame.
 Staple~\cite{bertinetto2016staple} combines color distribution information in a ridge regression framework to solve motion blur and illumination in tracking.
 All above correlation filter-based trackers treat each channel of features the same.
 CSR-DCF~\cite{Lukezic2017Discriminative} introduces the channel and spatial reliability into correlation filter framework.

 Secondly, we study the robust tracking methods with occlusion handling.~\cite{Li_2015_CVPR,li2017visual} are patch-based trackers, which are able to solve occlusion problem to some degree.
 However, they have inferior results on OTB-100 \cite{wu2015object} and VOT-2018\cite{VOT_TPAMI}. This is mainly due to that they pay too much attention to a small region for localizing the target and miss the global information. Besides, there are some keypoints-based trackers ~\cite{doi:10.1080/01691864.2015.1129360,hong2015multi}.
 They detect SIFT\cite{ng2003sift} keypoints in the search region, and then match these keypoints in the next frame to obtain target's position.
 Another popular strategy is based on computing the quality of response.
 MOSSE~\cite{5539960} estimates the tracking quality by computing the PSR~\cite{mahalanobis1987minimum} value of response map, where
 PSR~\cite{mahalanobis1987minimum} is used to capture the relationship between variance and the maximum value of response.
 ~\cite{DBLP:journals/corr/WangLH17} defines a quality function similar to PSR~\cite{mahalanobis1987minimum} while it is based on maximum and minimum value. 
~\cite{DBLP:journals/corr/WangLH17} employs APCE ~\cite{DBLP:journals/corr/WangLH17}in its updating process to deal with occlusions. The tracker stops updating when APCE~\cite{DBLP:journals/corr/WangLH17}value is smaller than a threshold. On the other hand,~\cite{bhat2018unveiling} proposes a response quality measurement function and proves its effectiveness.
 FCLT~\cite{lukevzivc2017fclt} introduces a model updating policy for two different kinds of filters to tackle occlusion problem.
 
 Finally, we review the methods to deal with scale variations. There are two main streams of solutions. DSST~\cite{danelljan2014accurate} and SAMF~\cite{Li2014A} enumerate a pyramid of samples to estimate the target scale variation by either sophisticated function or pooling. Alternatively, Li et al.~\cite{DBLP:journals/corr/abs-1712-05231} suggest to use phase-correlation in Log-polar coordinates to directly estimate the scale change.

\section{Correlation Filter Tracker with Occlusion Handling}
In this section, we present our proposed correlation filter-based tracking approach with effective occlusion handling. We choose the CSR-DCF~\cite{Lukezic2017Discriminative} as our base tracker, which is the winner of realtime sub-challenge in VOT2017\cite{VOT_TPAMI}. Then, we introduce our proposed tracking scheme. Finally, we present an effective scale estimation method by fusing the log-polar and pyramid scale estimator.
\subsection{CSR-DCF Tracker}

Due to its highly efficiency and promising performance, we formulate our proposed tracking approach under the framework of CSR-DCF~\cite{Lukezic2017Discriminative}, which introduces the channel and spatial reliability into the filter learning process.

Given a search region of size ${C}\times{R}$ , a set of $N_d$ feature channels $\textbf{u}=\{\textbf{u}_c\}_{c=1:N_d}$, where $\textbf{u}_c\in{\mathbb{R}^{C\times{R}}}$ are extracted. In CSR-DCF, the spatial reliability map adjusts the filter according to the part of object suitable for tracking, which results in the increment on accuracy and robustness. Moreover, spatial reliability is obtained by learning a binary mask $\textbf{m} \in \{0,1\}$ with the size of $C \times R$. The binary mask is calculated according to the percentage of foreground color histogram in the whole picture. Then, a set of $N_d$ learned correlation filters $\textbf{f}=\{\textbf{f}_d\}_{d=1:N_d}$ is multiplied by $\textbf{m}$, where $\textbf{f}_d \in{\mathbb{R}^{C\times{R}}}$. The discriminative filter $\textbf{f}$ is learned by minimizing the following Augmented Lagrange~\cite{boyd2011distributed}: 
\begin{equation}
                      \begin{split}
                      \mathcal{L}(\hat{\textbf{f}_c},\textbf{f},\hat{\textbf{I}}|m)=\Vert \textbf{diag}{\hat{(\textbf{u})}\overline{\hat{\textbf{f}_c}}}-\hat{\textbf{g}}\Vert ^{2} +  
\frac{\lambda}{2} \Vert \textbf{f}_m\Vert ^{2} + \\
[ {\hat{\textbf{I}}^H(\hat{\textbf{f}_c}-\hat{\textbf{f}_m}) + \overline{ \hat{\textbf{I}}^H(\hat{\textbf{f}_c}-\hat{\textbf{f}}_m}) ]+ \mu \Vert \hat{\textbf{f}}_c-\hat{\textbf{f}}_m\Vert ^{2} }   
\end{split}
\end{equation}
where $\hat{\textbf{I}}$ is a complex Lagrange multiplier. ${\textbf{g}}$ is the desired output, which is usually the two-dimensional Gauss distribution with the largest value in the center of output like a square bell. $\hat{(\cdot)}$ denotes Fourier transformed variable. $\overline{{(\cdot)}} $ represents complex conjugate. ${{(\cdot)^H}} $ denotes  transposition. $\textbf{f}_m=(\textbf{m}\odot \textbf{f})$ represents that $\textbf{f}_m$ is derived from the result of element-wise multiplication between the spatial mask $\textbf{m}$ and $\textbf{f}$. $\textbf{f}_c$ is a dual variable with $\textbf{f}_c - \textbf{f}_m \equiv 0$.  $\lambda$ and $\mu$ are the parameters for constraint items. 

As in~\cite{Lukezic2017Discriminative}, the solutions of the above minimization can be obtained through ADMM~\cite{boyd2011distributed} iterations between two closed-form solutions:
  \begin{equation}
  \hat{\textbf{f}}_c^{i+1}=(\hat{\textbf{u}} \odot \overline{\hat{\textbf{g}}}+(\mu \hat{\textbf{f}}_m^i-\hat{\textbf{I}}^i)) \odot ^{-1} (\overline{\hat{\textbf{u}}}\cdot{\hat{\textbf{u}}}+\mu ^i)         
  \end{equation}
  \begin{equation} \label{eqn2}
  \textbf{f}^{i+1} = \textbf{m} \odot \mathcal{F} ^{-1} [\hat{\textbf{I}}^i+\mu{^i} \hat{\textbf{f}}_c^{i+1}]/(\frac{\lambda}{2D} + {\mu}^i)
  \end{equation}
where $\mathcal{F}^{-1}(\cdot )$ denotes Inverse Fourier Transform. $\odot$ represents element-wise multiple, $\odot ^{-1}$ is element-wise divide. $D=C \cdot R$. After getting the result of $\textbf{f}^{i+1}$, CSR-DCF~\cite{Lukezic2017Discriminative} computes the filter response map as follows:
         \begin{equation} {  \textbf{r}=\sum_{d=1}^{N_d}(\textbf{u}_d * \textbf{f}_d)\cdot w_d } 
         \end{equation}
where $w_d$ is computed according to the discrete responses with respect to the different feature channels. The maximum response on the training region reflects the target localization. Then, the binary mask $\textbf{m}$ will be updated with the new search region. 
 
\subsection{Correlation Filter Tracking with Occlusion Handling}
Usually, two different correlation filter models are used in the effective occlusion handling scheme. As in~\cite{lukevzivc2017fclt}, our proposed tracker tries to learn the filter model $\textbf{h}_t$ at frame $t$ as below:
\begin{equation} \label{eq1} 
\textbf{h}_t=
\begin{cases}
\textbf{d}_t, { s_t>0 } \\
\textbf{f}_t, { s_t\leq 0} \\
\end{cases} 
\end{equation}
 where $\textbf{d}_t$ is an occlusion handling filter in case of special situations, i.e., occlusions and drifting. $\textbf{f}_t$ is a regular tracking filter computed from Eq.\ref{eqn2}. $s_t$ is the value to decide which filter will be used for tracking at current frame, which is calculated as follows:
 \begin{equation} \label{eq2} 
s_t= {{Q_t^{\textbf{f}_t}}} - {{Q_t^{\textbf{d}_t}}}  
\end{equation}
where $ Q_t^{\textbf{f}_t}$ denotes the quality measurement for the tracking filter $\textbf{f}_t$, and $Q_t^{\textbf{d}_t}$ is the quality measurement for $\textbf{d}_t$.

Let $ x \in {\mathbb{R}}^2$ denote the location in the response map. The tracking quality  $Q_t$ for the correlation filter-based tracker is usually measured by the response-map $r(x)$:
            \begin{equation}\label{eq3}
            Q_t={q(r(x))}\cdot{\max(r(x))}  
            \end{equation}
where ${q(r(x))}$ represents the measurement function. Typically, both PSR~\cite{mahalanobis1987minimum}
and Prediction Quality Measure~\cite{bhat2018unveiling} can be used as the measurement function.
     
  Unlike~\cite{bhat2018unveiling}, we propose a general quality measurement function as follows:   
            \begin{equation}\label{eq4}
q(r(x))=\min \limits_{x}\frac{( r(x^*)-r(x))^{\alpha}}{1-e^{-\beta \arrowvert{x-x^*} \arrowvert^2}}
\end{equation}
where $x^*$ is the location with the largest score. $\alpha \in \mathbb{R}$ is a parameter that adjusts the non-linearity of numerator in the above equation. 

Generally, $\alpha $ lies in the range of $\alpha \in [1,\infty ) $, which should not be less than 1. Or $\alpha -1 < 0$. It may cause that the derivative of numerator of Eq. \ref{eq4} including $( r(x^*)-r(x))^{\alpha-1}$  will decrease as $(r(x^*)-r(x))$ is increasing, not increase as Prediction Quality Measure~\cite{bhat2018unveiling} or ours function wishes. At this time, $q(r(x))$ will get its value where $x$ locates a much near place of $x^*$. Then it loses the property of taking trade-off between the accuracy and the robustness.

   In order to analyze the effectiveness of our proposed measurement function, we study how $q(r(x))$ changes when $x$ moves from the place nearby $x^*$ to the edge of response map. It is usually far away from the peak location $x^*$. When $x$ is very close to $ x^*$, the denominator of above equation is very close to zero, which is invalid for analysis.  Assume that $ q(r(x)) $ is twice continuously differentiable and have a positive domain. We denote the gradient and Hessian of $r$ at $x$ as $\nabla {r}$ and $\lambda_2\leq \lambda_1 \leq 0$, where $\lambda_1$ and $\lambda_2$ are eigenvalues of $\mathbb{H}{r(x^*)} $. According to Hospital Rule~\cite{lukevzivc2017fclt}, we can get the inequality as below: 
\begin{equation}\label{eq5}
q(r(x)) \leq \frac{C_1\cdot{(r(x^*)-r(x))^{\alpha -1}}\vert \lambda_1\vert}{\beta}   
\end{equation}
Note that $C_1=2\alpha$ is a constant value when $\alpha$ is a fixed value. Usually, $r(x^*)-r(x)$  is no more than 2 after normalizing $r(x)$. Then, the following inequality can be obtained:
            \begin{equation}\label{6}
q(r(x)) \leq C_2 \frac{\vert \lambda_1 \vert}{\beta}        
\end{equation}
where $C_2\leq 2C_1$ has the a upper boundary. Then, we can see that the eigenvalue $\lambda_1$ is able to bound the value of $q(r(x))$. Note that the eigenvalue $\vert \lambda_1 \vert$ represents the minimum curvature of the response distribution $r(x)$ at the peak $x^*$. Thus, it is a kind of representation on the sharpness of response peak. The large value of $q(r(x))$ ensures that the peak is good enough for localization. These parameters $\alpha $ and $\beta$ adjust the trade-off between the robustness and accuracy of our proposed tracker.

Then, $x$ moves to the place which locates at the middle point between $x^*$ and the edge of response map. As $q(r(x))$ is the minimal value of one fractional, we obtain its derivative and make it to be vanished:
\begin{equation}\label{eq123}
\frac{d(\frac{( r(x^*)-r(x))^{\alpha}}{1-e^{-\beta \vert{x-x^*}\vert^2}})}{dx} = 0
\end{equation}
Then:
\begin{equation}
(-\frac{d(r(x))}{dx})(1-e^{-\beta \vert{x-x^*}\vert^2})-(\beta{\vert{x^*-x}\vert}^2\cdot{\frac{d(\vert{x^*-x}\vert^2)}{dx}}(r(x^*)-r(x))^\alpha) = 0
\end{equation}Let's look at the left side of above equation and see what range of value it can get. After analysis, we can see that it can get one value domain contains 0. 
As there are several local maximum or minimum values in $r(x)$. So above equation can be simplified at these places as: 
\begin{equation}
left = -(\beta{\vert{x^*-x}\vert}^2\cdot{\frac{d(\vert{x^*-x}\vert^2)}{dx}})\cdot B 
\end{equation}where $B=(r(x^*)-r(x))^\alpha$ is one positive constant. The right hand of above equation can be a negative value or a positive value. Meanwhile, it can get both negative and positive values because there are local maximum or minimum values distributed uniformly around the $x^*$. So, the right hand of above equation can obtain negative and positive value at the same time.  As $q(r(x))$
is twice continuously differentiable, its derivative has smoothness,  Therefore, right hand of above equation has value belongs to $[-T,O]$ which contains 0, where $T$ and $O$ are positive value.  As right hand of above equation can get value of 0. Then it can be derived that Eq.\ref{eq123} can be met. 
So, we can say that $q(r(x))$ can get its value in the middle place at the axis where $x$ moves from $x$ to the edge of the response.

 When $t$ moves to a place far away from the $t^*$, where the denominator is close to one. So, we can obtain the following about-equation:
\begin{equation}\label{7}
q(r(x)) \approx  {(r(x^*-r(x)))^{\alpha}}    
\end{equation}
That is to say, $q(r(x)) \propto ({r(x^*)-r(x)})^\alpha $. Then the response score quality is bounded by the immediate neighborhood of the prediction $x^*$. Thus, the large score of $q(r(x))$ ensures that there is no disturb to make the prediction uncertain. On the contrary, if there are other peaks with a similar score $r(x^*) \approx r(x)$, then the score $ q(r(x)) $ is low.

\begin{figure*}[htbp]
\centering  
\subfloat[]{
\begin{minipage}[t]{0.20\linewidth}  
 \includegraphics[width=1\textwidth] {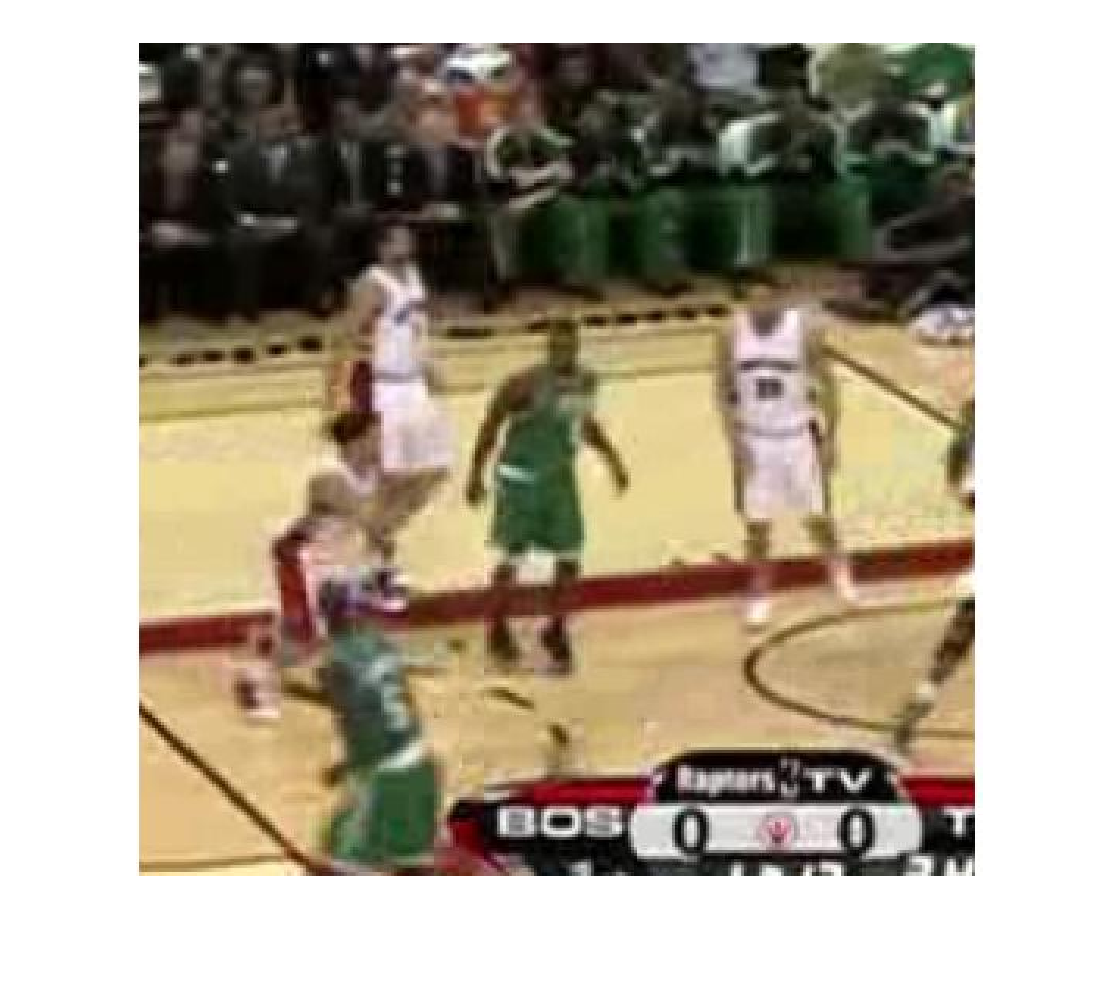}
\end{minipage}%
}
\hfill
\centering
\subfloat[]{
\begin{minipage}[t]{0.25\linewidth}  
\centering  
   \begin{overpic}[width=1\textwidth]{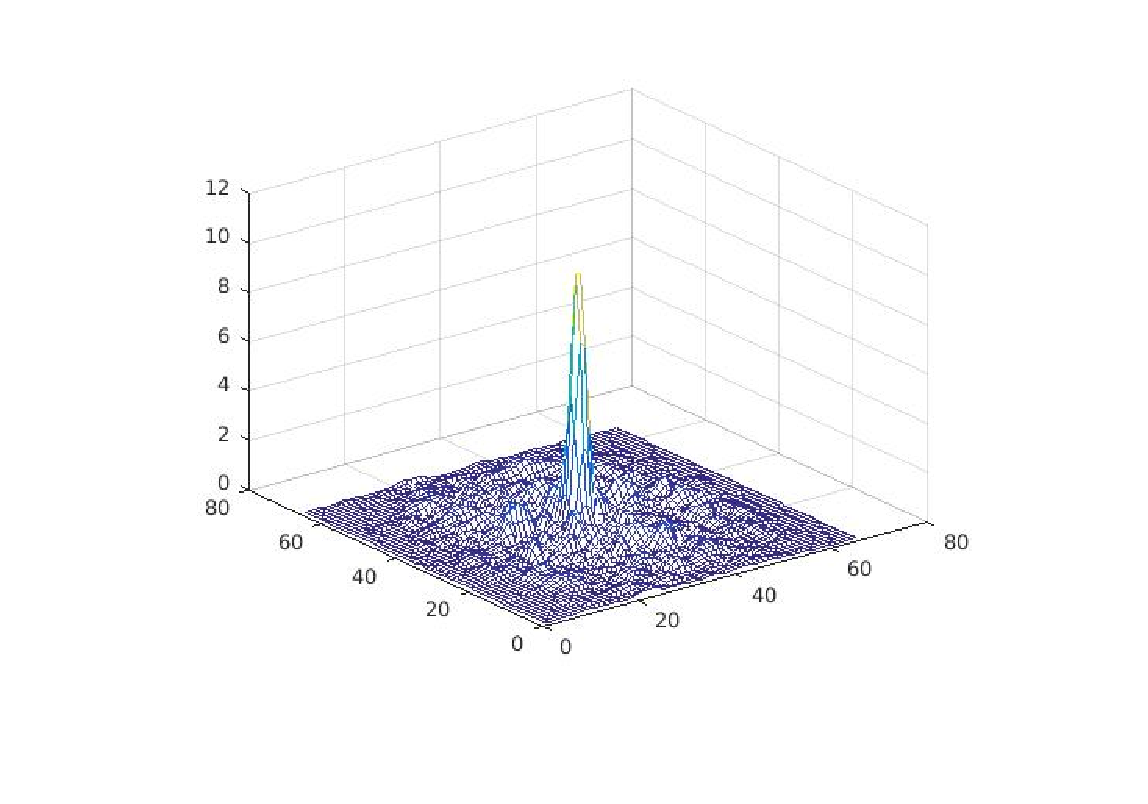}
\put(15,60){ \color{red}{${q(r(1))=7.9\cdot 10^{-5}}$}}
\end{overpic}
\end{minipage}  
}
\hfill
\centering  
\subfloat[]{
\begin{minipage}[t]{0.20\linewidth}  
 \includegraphics[width=1\textwidth] {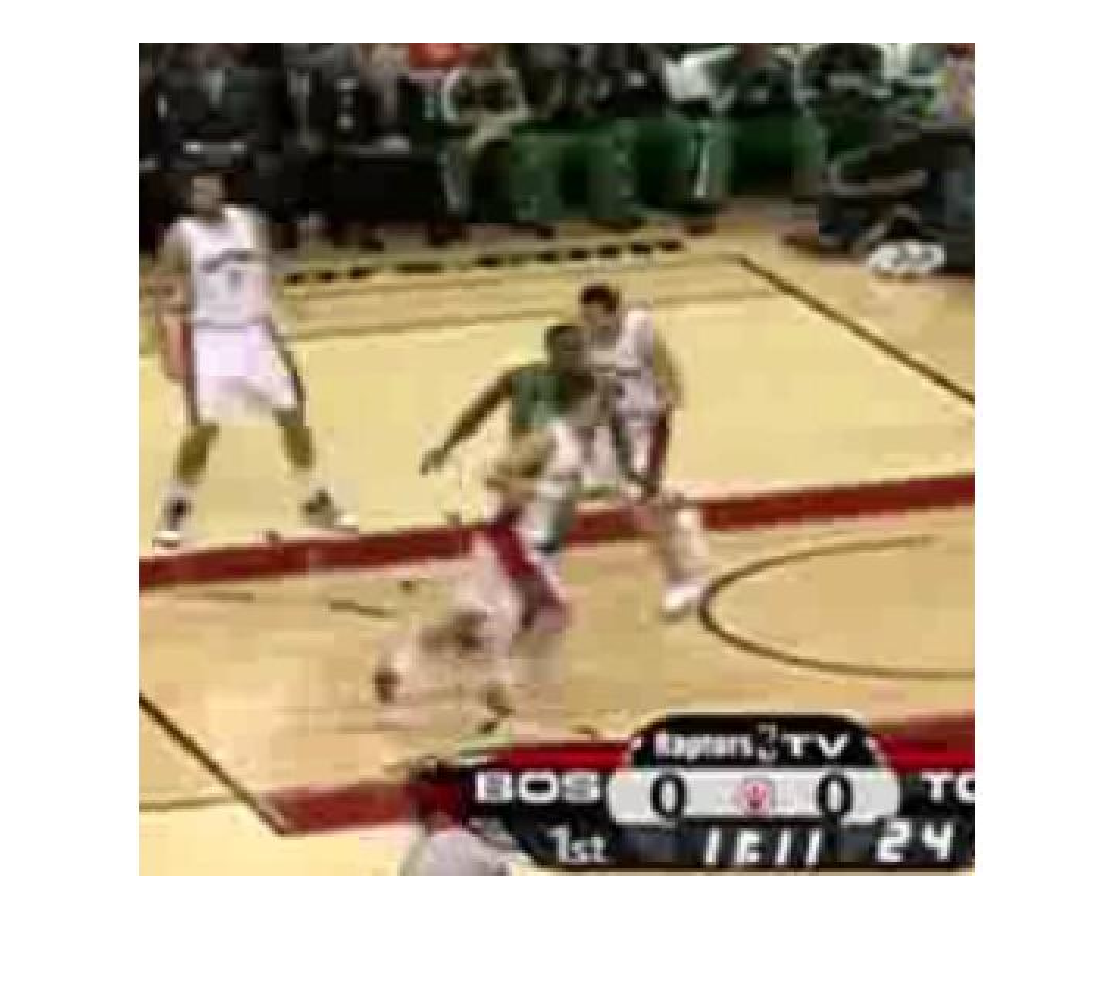}
\end{minipage}%
}
\hfill
\centering
\subfloat[]{
\begin{minipage}[t]{0.25\linewidth}  
  \begin{overpic}[width=1\textwidth]{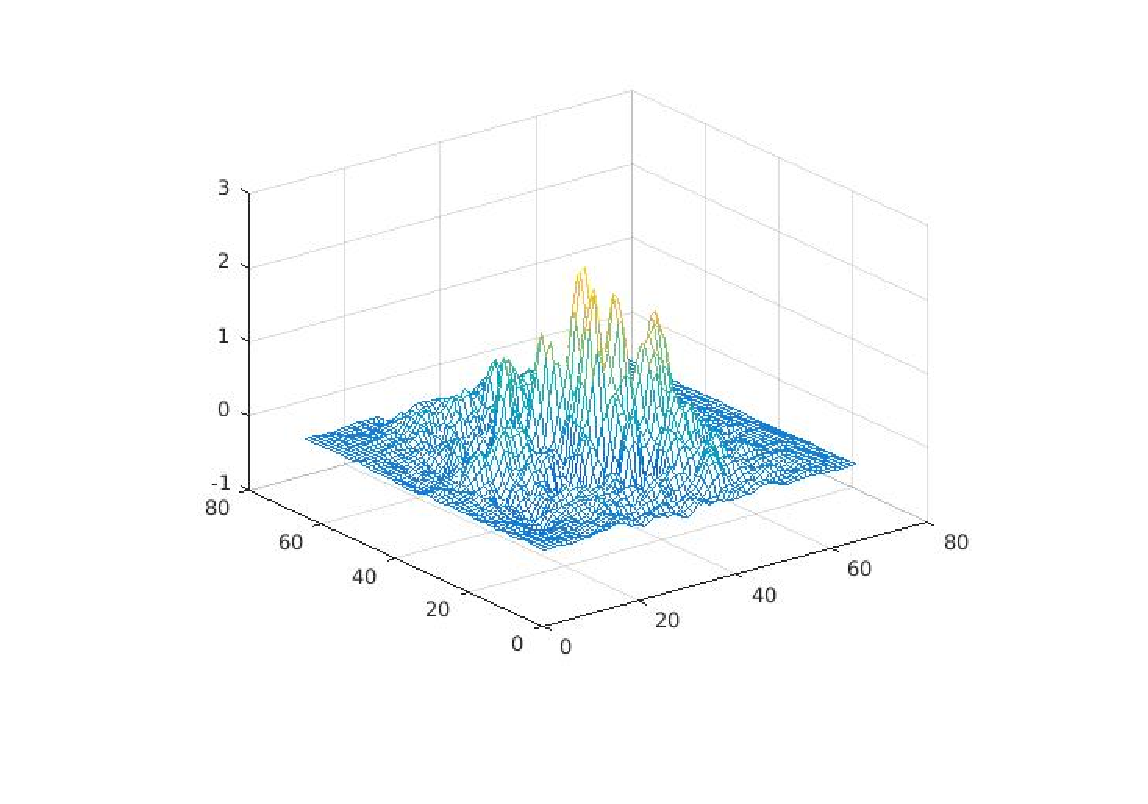}
\put(15,60){ \color{red}{${q(r(19))=3.9\cdot 10^{-7}}$}}
\end{overpic}
\end{minipage}  
}
\caption{(a) Example frame with the target not being occluded, and the corresponding response map has single sharp peak in (b); (c) Example frame with occluded and deformed target, whose response map has multiple peaks.}  
\label{fig1}
\end{figure*}

From the above discussion, $q(r(x))$ obtains its value with $x$ at a middle place on the axis from $x^*$ to the edge of response map. Therefore, it is clear that $q(r(x))$ takes into account robustness and accuracy at the same time. The non-linearity of numerator of the measurement function in ~\cite{bhat2018unveiling} is smaller than that of the denominator. At the meantime, both of the numerator and denominator are the monotone increasing function about ${r(x^*)-r(x)}$. Therefore, denominator with more non-linearity changes quicker than numerator, which is likely cause $q(r(x))$ that the value very quick when $x$ move from the axis from $x^*$ to the edge of the response. It is contradictory for our requirement to get $q(r(x))$ with $x$ in the middle place for good trade-off between the accuracy and the robustness. So, We take the numerator as non-linear function to fix it by increasing the numerator's power degree. As shown in Fig.~\ref{fig1}, when target appearance is clear and non-occluded, the $q(r(x))$ is high. However, $q(r(x))$ is much lower when occlusion happens to the target.
   
           
\subsection{Model updating}
\subsubsection{Tracking filter}

   Our tracker employs the filter learned from CSR-DCF~\cite{Lukezic2017Discriminative} to track the target in case of few occlusion or small deformations on the target's appearance. Moreover, the target position is estimated as the location at the maximum of  correlation response between the tracking filter and the features extracted from the search region. We update the tracking filter $\textbf{f}_t$ with the simple updating strategy by linear interpolation with the filter model from the previous frame:
            \begin{equation}\label{eq8}
\textbf{f}_{t}= (1-\eta)\textbf{f}_{t-1}+\eta \widetilde {\textbf{f}}_{t-1}   
\end{equation}
        where $\widetilde {\textbf{f}}_{t-1}$ is the correlation filter learned in the previous frame, and $\eta$ is the updating factor. ${\textbf{f}}_{t-1}$ is a correlation filter from the previous number $t-1$ frames. 

\subsubsection{Occlusion handling filter}
   When the target is occluded, the tracking capability of learned filter will be degraded. It is hard to handle the occlusions, especially when the target appearance changes quickly. The filter will have few time to adapt to the rapid changes. If our tracker uses these patches with part-reliable or deformed appearance, the learned filter cannot locate the target accurately. When the target's appearance changes, its response map should varies accordingly. Therefore, this problem can be solved by check the response distribution from the tracking filter. Then, our proposed tracker is able to detect the  occlusions by activating the occlusion handling scheme $\textbf{d}_t$, In this paper, occlusion handling filter can be built with the combination of tracking filter and its original state: 
\begin{equation}\label{9}
\textbf{d}_{t}=(1-\xi_{t}) \textbf{f}_{0} + \xi_{t}\textbf{f}_{t}         
\end{equation}
 where the weight $\xi_{t} = e^{-\alpha_{d}(\Delta t)^2}$  depends on the factor $\alpha_{d}$ and the number of frames $\Delta_{t
}$. $\Delta_{t}$ is the number of the frames since the last confidently estimated position of target. Since occlusion may accumulate the changes, the target's appearance becomes more and more different from its initial state. Therefore, we set $\xi_{t}$ to decrease as $\Delta_{t}$ increases. Once the occlusion handling process is finished and the normal tracking filter is employed, we set $\Delta_t$ to be zero and $\textbf{d}_t=\textbf{f}_t$. 
 
\subsubsection{Occlusion Detection}
 We denote $\bar{Q}$ as the average localization quality over the recent $N_q$ frames. Inspired by ~\cite{bhat2018unveiling}, occlusion can be reflected in a significant drop in value of $Q_t$. It can be detected when the ratio between $\bar{Q}$ and $Q_t$ is larger than a given threshold $\phi$. is one method  which is fast but suffers from the random errors.
 
 DSST~\cite{danelljan2014accurate} has an advantage of a large set of scales, but it runs not fast
\begin{equation}\label{eq10}
\frac{\bar{Q}}{Q_t}>\phi  
\end{equation}        
         Our tracker is initialized in the first frame and the learned initialization filter model $\textbf{f}_0$  is stored. In the remaining frames, two visual models are maintained at different time for target localization. The tracking model $\textbf{f}_t$  and the occlusion handling model $\textbf{d}_t$. In the first few frames, the response quality scores are good. A tracking processing at frame $t$ starts with the target's position $x_{t-1}$  from the last frame as well as occlusion detection score $\frac{\bar{Q}_{t-1}}{Q_{t-1}} $ and the mean $\overline{Q}$ over recent ${N_q} $ confidently tracked frames. The extracted region centered at location $ x_{t-1} $ in the current image, and the correlation response is computed by convolving the feature of region with $\textbf{f}_{t-1}$. Position $ x_{t}^{\textbf{f}}$ and localization quality $ Q_{t}^{\textbf{f}}$  are estimated from the correlation response $r_t^{\textbf{f}} $.

        When occlusion occurs at the first time, $\textbf{d}_t $ will be activated. Then, if  $\frac{\overline Q^{\textbf{f}}}{\bar{Q_t^{\textbf{f}}}} $and $\frac{\overline Q^{\textbf{d}}}{\bar{Q_t^{\textbf{d}}}}$ are both higher than $\phi$, the position of target and these two learned models will stop updating. Otherwise, The position is decided by the model with larger ${Q_t}$. $\textbf{f}_t$ and $\textbf{d}_t$ update by their own rules. The whole approach is summarized in Algorithm~\ref{alg1}. 
         
\subsection{Fusion approach with scale estimation}
To deal with scale changes, we present a fusion approach to take advantage of both pyramid sampling method and the phase correlation in log-polar coordinates. As DSST~\cite{danelljan2014accurate} enjoys the merit of sampling a large set of scales, which leads to stable results. On the other hand, the phase correlation may obtain more accurate target size while suffering the issues of random failures. Therefore, we propose to combine the estimate result from both methods to take into account of both robustness and accuracy. We represent $S$ as the scale our tracker want to get: 
\begin{equation}\label{15}
S=\theta S_d +(1-\theta )S_p      
\end{equation}
          where $S_d$ is the size estimation using DSST~\cite{danelljan2014accurate},and $S_p$ is the results coming from log-polar~\cite{DBLP:journals/corr/abs-1712-05231}.
\begin{algorithm}[htb] 
\caption{ Framework of Our tracker} 
\label{alg1} 
\begin{algorithmic}[1] 
\REQUIRE ~~\\ 
Given target's position $x_0$;Learned normal situation model $\textbf{f}_0$;
$\Delta_{t} = 0$, $D_0$.
\ENSURE ~~\\ %
Updated position of target, $x_t$;Updated normal situation model, $\textbf{f}_t$;\\
Activation and updating of special situation model, $\textbf{d}_t$ and $\Delta_{t}$;
\STATE for every $t$ in sequence:
\IF{$\frac{\overline{Q}^{\textbf{f}}}{Q_t^{\textbf{f}}}<\phi$}
\STATE Update $\textbf{f}_t$ according to eq. \ref{eq8}
\ELSE
\STATE $\textbf{f}_t=\textbf{f}_{t-1}$
\ENDIF
\IF{$\frac{\overline{Q}^{\textbf{d}}}{Q_t^{\textbf{d}}}>=\phi$ \&\& $ \frac{\overline{Q}^{\textbf{f}}}{Q_t^{\textbf{f}}}>=\phi $ }
\STATE $\textbf{h}_t = \textbf{h}_{t-1}$
\ELSE
    \IF{$\frac{\overline{Q}^{d}_t}{Q_t^{d}}<\phi$}
         \IF{$\Delta_t==0$} 
           \STATE Activate $D_t$: $\Delta_t =1$
         \ELSE 
           \STATE Update $D_t$(eq.\ref{eq10}): $\Delta_{t} +=1$
        \ENDIF 
           \IF{$\frac{\overline{Q}^{\textbf{f}}}{Q_t^{\textbf{f}}}<\phi$}
          \STATE Update $H$ according to eq.\ref{eq1}    
        \ELSE
         \STATE $\textbf{h}_t=\textbf{d}_t$ 
        \ENDIF
    \ELSE
    \STATE Re-activate $D_t$: $\Delta_{t} =0$(eq. \ref{eq10})
    \STATE $\textbf{h}_t=\textbf{f}_t$
    \ENDIF
\ENDIF
\STATE Update $x_t$
\end{algorithmic}
\end{algorithm}
                                                           
\section{Experiments}
\label{sec:blind}
We conduct three experiments to evaluate the efficacy of our tracker. Firstly, we implemented three trackers with different response quality measurement function: PSR~\cite{mahalanobis1987minimum},  APCE~\cite{DBLP:journals/corr/WangLH17}, ours. We compare them with each other on VOT-2018~\cite{VOT_TPAMI} real-time sub-challenge. Secondly, we set $\alpha $ with different values and evaluate it on real-time sub-challenge to evaluate the stability of our proposed tracker. Additionally, we evaluate our tracker against the real-time state-of-the-art trackers on VOT-2018 real-time sub-challenge and main challenge ~\cite{VOT_TPAMI}. Finally, we evaluate our tracker against the real-time correlation filter-based trackers on OTB100~\cite{wu2015object}.

\subsection{Parameters setting and Experimental methodology}

Our tracker is implemented in C++ using OpenCV~\cite{opencv_library}. 
We used HOG~\cite{felzenszwalb2010object} and Color-naming~\cite{danelljan2014adaptive} features to learn the tracking filters and the occlusion handling filters. The cell size of HOG~\cite{felzenszwalb2010object} is 4 $\times$ 4 and the orientation
bin number of HOG~\cite{felzenszwalb2010object} is 9. Based on the discussion in the section of formula, we select the parameter used to tradeoff the robustness and accuracy of our tracker $\beta = 8$. The parameter for filter mixing in detector contraction is set to $\alpha_d=0.05$. The uncertainty threshold in Eq.~\ref{eq10} is set to $\phi = 45$. Recent frames number for computing $\overline{Q}$ is set to $N_q=100$. For the DSST~\cite{danelljan2014accurate} parameters, scale-increasing $l_r$ is 1.05.  The combination coefficient between DSST~\cite{danelljan2014accurate} and log-polar~\cite{DBLP:journals/corr/abs-1712-05231} $\theta$ is set to 0.2. Our experiments were conducted on a PC with Intel Core i7 4790K, 4.0GHz. 
            \begin{table} 
            	\vspace{-0.3in}
            	\begin{center}
            		\caption{Comparison experiments on quality measurement functions}
            		\label{table:headings}
            		\label{table1}
            		\begin{tabular}{l|c|c|c}
            			\hline\noalign{\smallskip}
            			Measurement methods & PSR ~\cite{mahalanobis1987minimum} & APCE~\cite{DBLP:journals/corr/WangLH17}&Ours\\
            			\noalign{\smallskip}
            			\hline
            			\noalign{\smallskip}
            			Results on VOT2018 realtime ~\cite{VOT_TPAMI} & 0.2228& 0.2230& \textbf{0.2320}\\
            			\hline
            		\end{tabular}
            	\end{center}
            	\vspace{-0.5in}
            \end{table}

\subsection{Comparison experiments with other quality measurements}
            In order to evaluate the effectiveness of our proposed quality measurement function, we compare the different quality measurement functions with the base CSR-DCF tracker~\cite{Lukezic2017Discriminative} on VOT 2018 realtime sub challenge~\cite{VOT_TPAMI}. Table \ref{table1} shows the results of them. Our tracker($0.2320$) outperforms other methods at least $4.0\%$.
            
            Besides, we vary $\alpha$ in the formula of $q(r(x))$ to see how its performance changes. When $\alpha$ equals to one, the proposed quality measurement function is degenerated to the one in~\cite{bhat2018unveiling}. From table \ref{table2}, it can be seen that our tracker obtains better performance than ~\cite{bhat2018unveiling}. When $\alpha = 2$, the proposed method gets the best result. For the larger $\alpha$, the results tend to be weaker. Note that we set $\alpha$ as two in the above experiments. It is worthy of noting our proposed generalized quality measurement function performs better than ~\cite{bhat2018unveiling} around $5.0\%$ ($0.2209 \rightarrow  0.2320$).

\begin{table} 
\begin{center}
\caption{Results on VOT 2018 real-time~\cite{VOT_TPAMI} for different {\it $\alpha$} in $q(r(x))$}
\label{table2}
\begin{tabular}{lllllllll}
\hline\noalign{\smallskip}
$\alpha$ &{   \vline} & 1 &1.5& 2 &2.5&3.5&5&9\\
\noalign{\smallskip}
\hline
\noalign{\smallskip}
EAO & {   \vline} & 0.2209&0.2210&\textbf{0.2320} &\textbf{0.2273}&0.2267&0.2244&0.2224\\
\hline
\end{tabular}
\end{center}
\vspace{-0.3in}
\end{table}

\subsection{Comparison with the state-of-the-art real-time trackers }
\subsubsection{VOT 2018 real-time sub-challenge}
We name our proposed tracker as "Channels weighted and spatial-related Tracker with Effective response-map Measurement" (CSTEM), which is same as our submitted entry in the challenge. 
             
\begin{figure*}
\begin{minipage}[t]{0.5\linewidth}  
\centering  
  \includegraphics[width=1\textwidth]{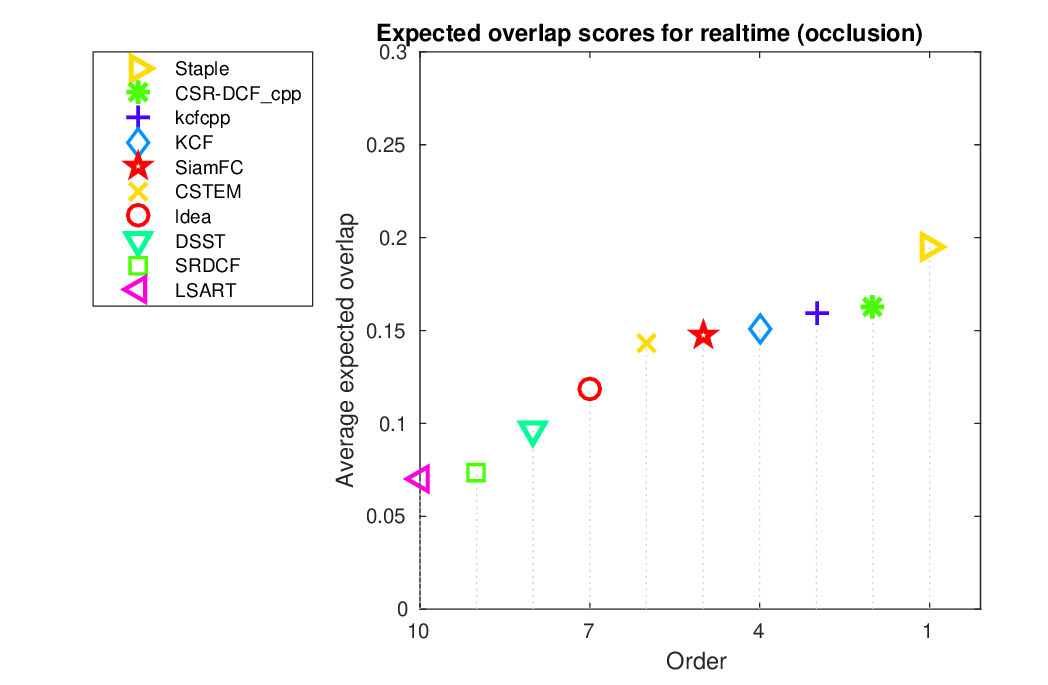}  
\label{fig:side:b}  
\end{minipage}  
\begin{minipage}[t]{0.5\linewidth}  
\centering  
  \includegraphics[width=1\textwidth]{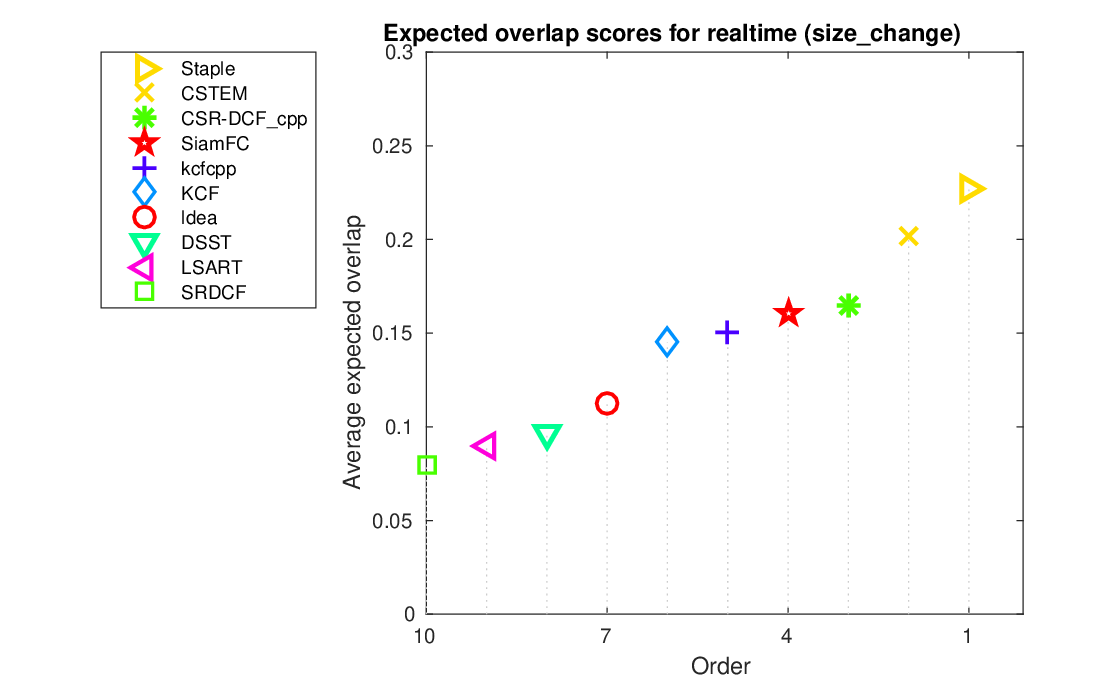}  
\label{fig:side:c}  
\end{minipage} 
\caption{ Scores on aspects of occlusion and size\_change of real-time trackers on VOT2018 ~\cite{VOT_TPAMI}.} 
\label{fig2}
\end{figure*}

The VOT 2018 real-time sub-challenge~\cite{VOT_TPAMI} dataset contains  60 videos, tracking performance is evaluated both in terms of accuracy (average overlap during successful tracking) and robustness (failure rate). Table \ref{table3} shows the results from the best correlation filter-based trackers. The last column presents the speed of each tracker. We calculated their speed according to the time data produced by the toolkit of VOT 2018~\cite{VOT_TPAMI}. All of them run more than 25 fps, which is the minimal requirement of real-time. Our tracker ranks moderate in terms of speed. On the other hand, for the EAO score, we can see the result of our tracker has the best result on VOT-2018 real-time sub-challenge~\cite{VOT_TPAMI}. In table \ref{table3}, the tracker KCF\_cpp, CSRDCF\_cpp are the C++ implementation for KCF~\cite{henriques2015high} and CSR-DCF~\cite{Lukezic2017Discriminative}, respectively. As shown in table \ref{table1}, our tracker achieves the best result on VOT2018 ~\cite{VOT_TPAMI} sub-challenge, which outperforms the winner tracker CSRDCF\_cpp of last year real-time sub-challenge by near 10\%($0.2320\rightarrow 0.2120$). Besides, Figure \ref{fig2} obtains the tracking accuracy and robustness rank of several trackers on dealing with occlusions and scale change problem. Our tracker achieves better results than that of the base CSR-DCF~\cite{Lukezic2017Discriminative} tracker. 

\begin{table}
	\vspace{-0.3in}
	\centering
	\centering
	\caption{Results of real-time trackers on VOT2018 realtime challenge}  
	\begin{tabular}{|c|c|c|}
		\hline
		\textbf{Trackers} & \textbf{Overall}& \textbf{Speed(fps)}\\\hline
		{CSTEM(proposed)} & {\textbf{ 0.2320}} & {41}\\\hline
		{CSR-DCR\_cpp~\cite{Lukezic2017Discriminative}} & {\textbf{ 0.2120}} & {49}\\\hline
		{Staple~\cite{bertinetto2016staple}} & 0.1696& {45}\\\hline
		{KCF\_cpp~\cite{henriques2015high}} & 0.1356& {92} \\\hline
		{KCF~\cite{henriques2015high}} & 0.1336 &{73}\\\hline
		{LDES~\cite{li2017visual}} & 0.1130 &{38}\\\hline
		{DSST~\cite{danelljan2014accurate}} & 0.0774 & {28} \\\hline
	\end{tabular}
	\label{table3}
	\vspace{-0.5in}
\end{table}

 \subsubsection{VOT 2018 main challenge}
 We evaluate our tracker on VOT 2018 main challenge~\cite{VOT_TPAMI} and make the comparison with real-time trackers. Our tracker obtains the best result EAO in the group of real-time trackers. Our tracker outperforms other trackers at least 9.1\% ($0.2294\rightarrow 0.2103 $). In summary, our tracker obtains the state-of-the-art performance in the group of real-time correlation-filter based trackers on VOT 2018 realtime and main challenge~\cite{VOT_TPAMI}.

\begin{table}
\label{table4}
\centering
\caption{Results of trackers on VOT-2018 main challenge~\cite{VOT_TPAMI}.}
\begin{tabular}{|c|c|c|c|c|c|c|c|}
\hline
\multirow{2}{*}{\textbf{}} 
\textbf{Trackers} &{CSTEM(proposed)} &{CSR-DCF\_cpp~\cite{Lukezic2017Discriminative}}&{Staple~\cite{bertinetto2016staple}}&{KCF~\cite{henriques2015high}}&{LDES~\cite{li2017visual}}&{DSST~\cite{danelljan2014accurate} }\\\hline
 \textbf{EAO} &{\textbf{0.2294}}& {\textbf{0.2103}}&{0.1694}&{0.1349 }&{0.1113}&{ 0.0788 }\\\hline
\end{tabular}
\end{table}
           
 \subsubsection{OTB100}
   The OTB100~\cite{wu2015object} benchmark contains 100 sequences with various target appearance changes in sequences. We plot top-ranking realtime correlation filter-based trackers on OTB100 benchmark~\cite{wu2015object} i.e. KCF~\cite{henriques2015high}, SAMF~\cite{Li2014A}, DSST~\cite{danelljan2014accurate}, SRDCF~\cite{danelljan2015learning}, BACF~\cite{galoogahi2017learning} . 
   
   \begin{figure*}
   	\includegraphics[width=0.5\textwidth]{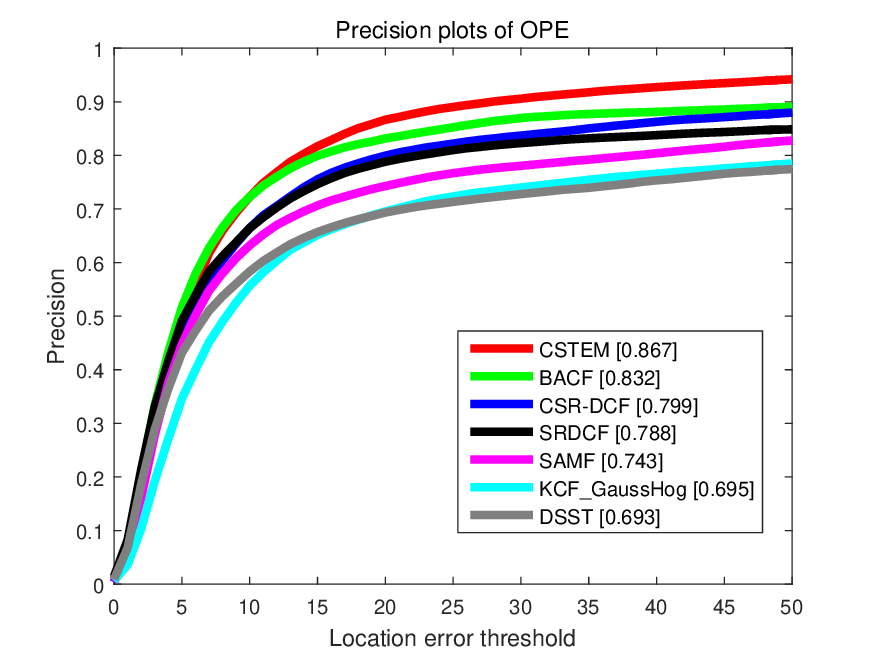}
   	\includegraphics[width=0.5\textwidth]{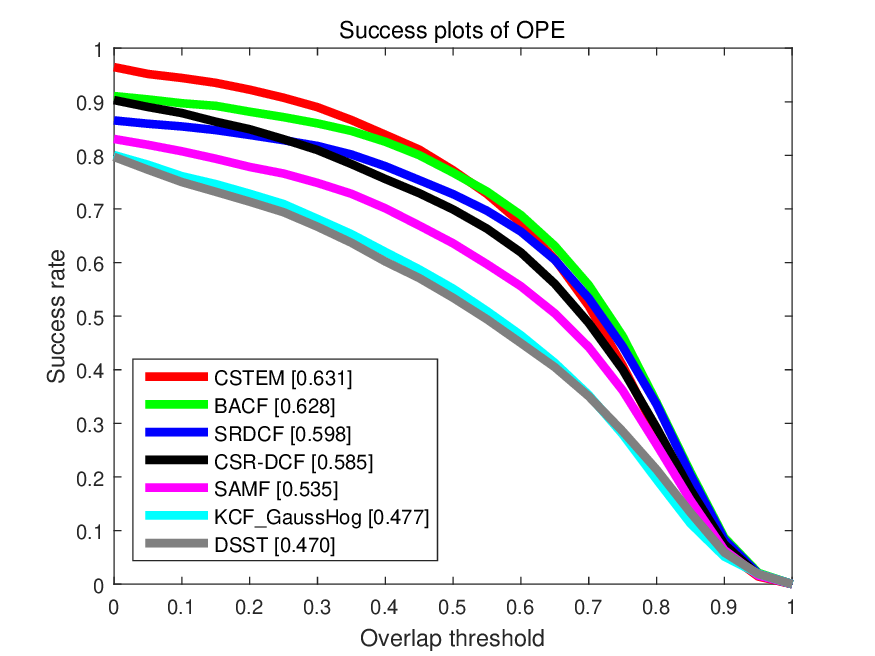}
   	\caption{Common real-time correlation filter-based trackers results on OTB100\cite{wu2015object}}
   	\label{fig3}
   \end{figure*} 
    As shown in Figure~\ref{fig3}, it can be observed that our proposed approach is the best tracker, which has improved more than 5 percent points than the based tracker (CSRDCF~\cite{Lukezic2017Discriminative}) from 58.5\% to 63.1\% in AUC score.

\begin{figure*}
	\centering
	\includegraphics[width=0.48\textwidth]{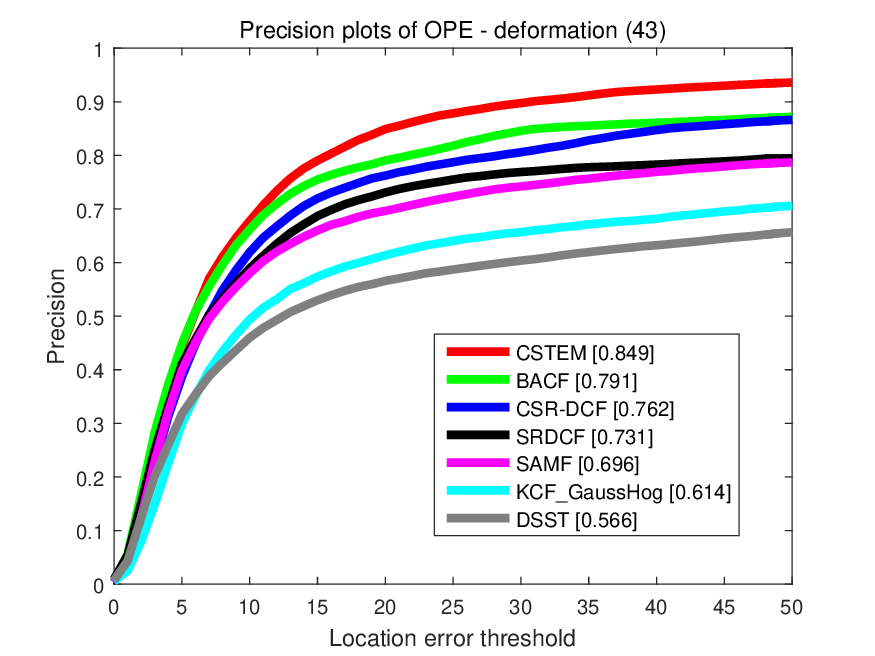}
	\includegraphics[width=0.48\textwidth]{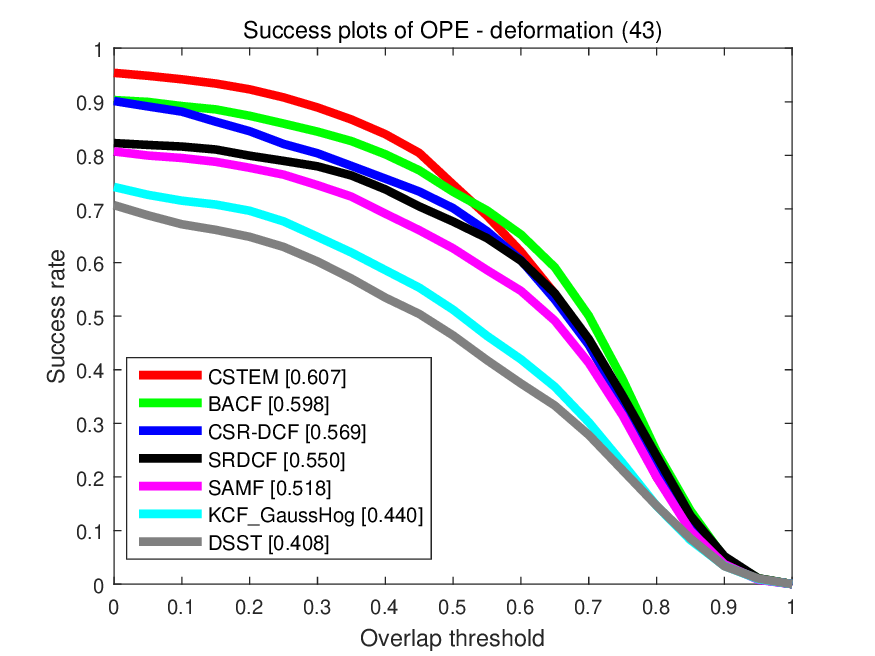}
	\includegraphics[width=0.48\textwidth]{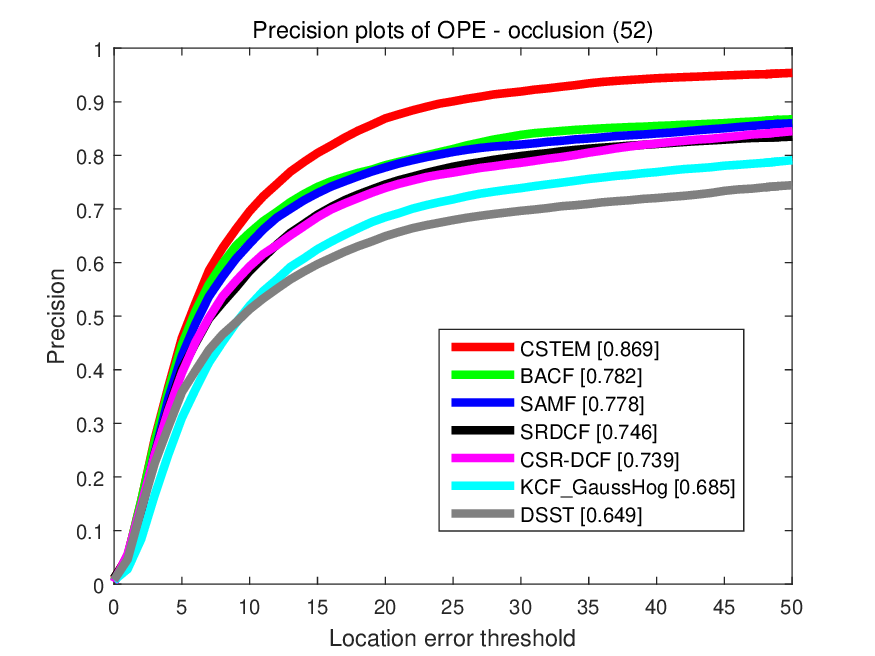}
	\includegraphics[width=0.48\textwidth]{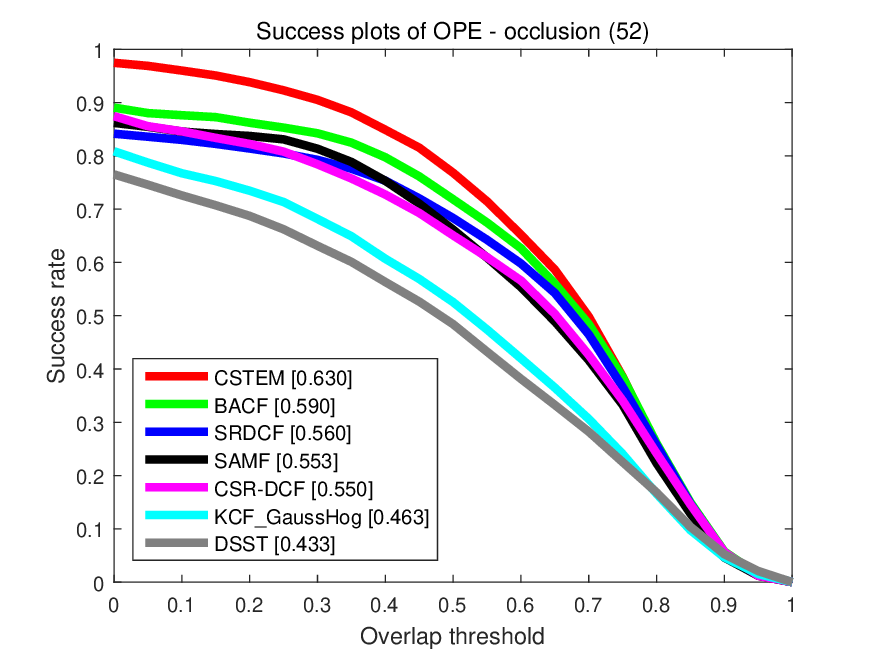}
	\includegraphics[width=0.48\textwidth]{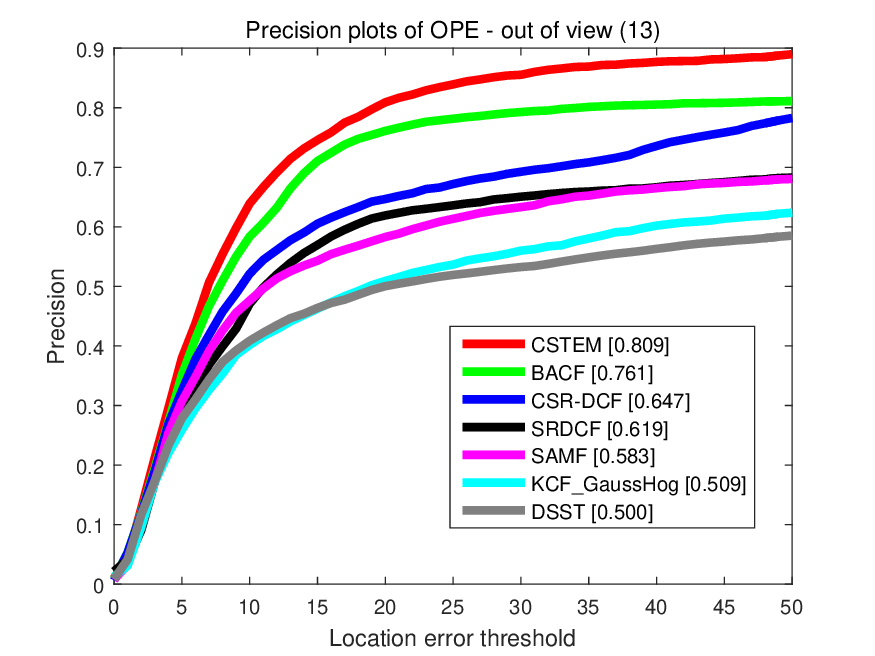}
	\includegraphics[width=0.48\textwidth]{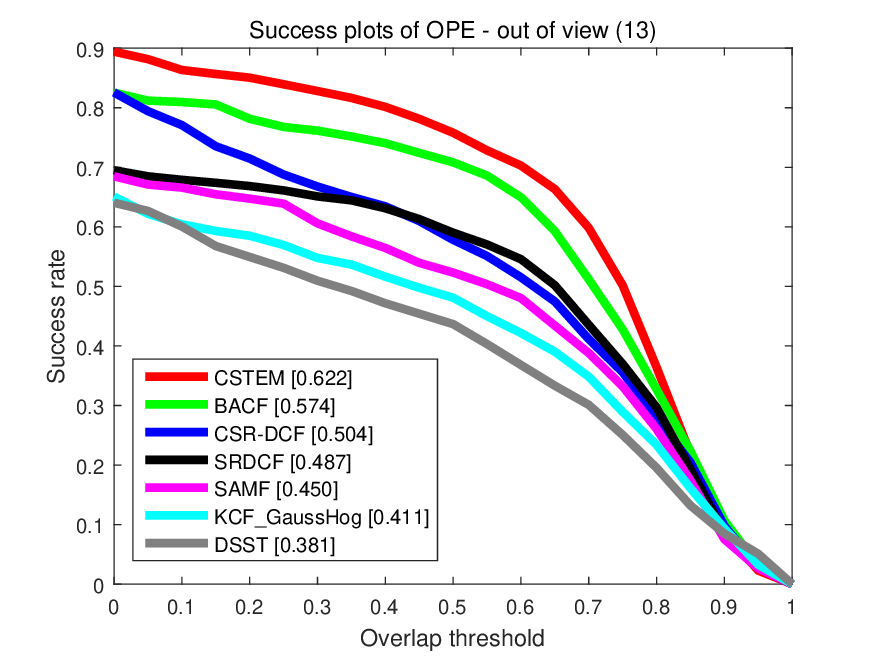}
	\includegraphics[width=0.48\textwidth]{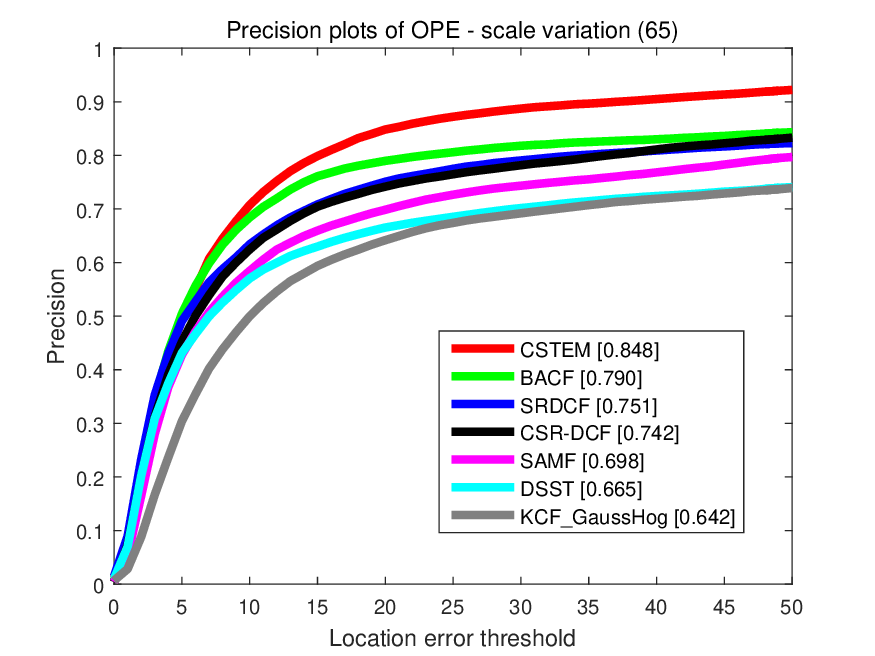}
	\includegraphics[width=0.48\textwidth]{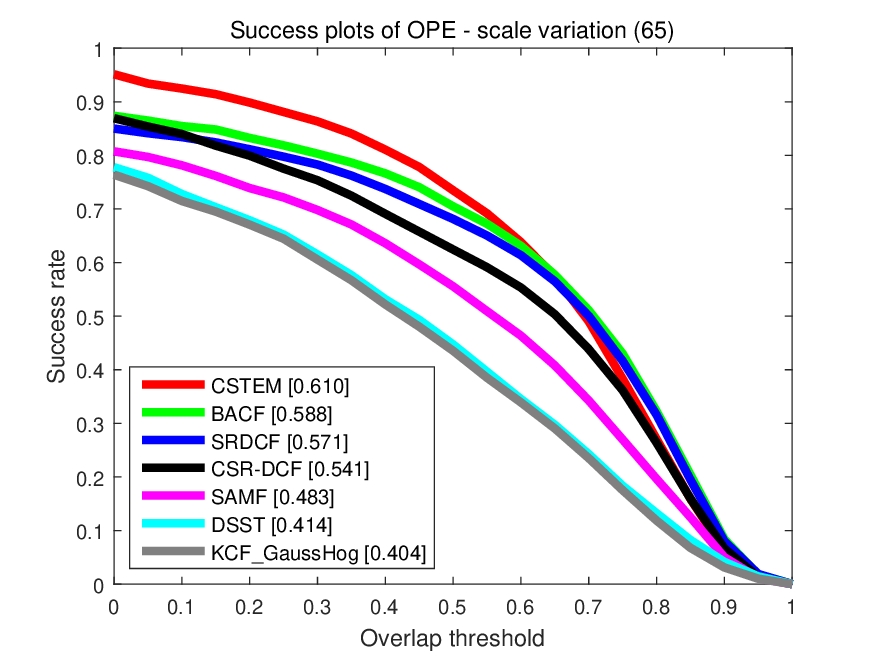}
	\caption{Common real-time correlation filter-based trackers' attributes results on OTB100\cite{wu2015object}}
	\label{fig4}
\end{figure*}
   Besides, our tracker outperforms against other trackers as for the some specific problems. From the first row of Fig.~\ref{fig4}, our tracker obtains the best result on handling deformation. From the second row, it is clear that our tracker outperforms other trackers in a big deal. The right picture shows the AUC score of all trackers, our tracker gets a score of 63.0\% which has 6.8\% improvement than the second one with a score of 59.0\%. This result proves the effectiveness of our tracker's on handling occlusion. From the third rows, we can see that our tracker has promising performance in dealing with out-view. The last row shows that our novel method of combining DSST~\cite{danelljan2014accurate} and phase correlation in log-polar coordinates~\cite{DBLP:journals/corr/abs-1712-05231} is a successful strategy to deal with scale variation problem.

\section{Conclusion}
We have proposed a novel correlation filter-based tracking scheme with effective occlusion handling, which is able to choose the specific filter model according to different scenarios. A sophisticated strategy has been employed to judge whether occlusions occurs through an effective measurement function to evaluate the quality of response. Moreover, different correlation filter models are used to track the targets with occlusions or not, respectively. Additionally, we take advantage of both log-polar method and pyramid-like approach to estimate the best scale of target. We have evaluated our proposed approach on VOT2018 real-time sub-challenge benchmark and OTB100 dataset, whose experimental result shows that the proposed tracker is capable of handling occlusions and alleviating the drifting issues.

\bibliographystyle{splncs}
\bibliography{egbib}
\end{document}